%% file: main.tex
\documentclass[preprint,12pt]{elsarticle}

\usepackage{amssymb}
\usepackage{amsmath}
\usepackage{url}

\usepackage{lineno}

\usepackage{graphicx,tikz,xcolor}

\usetikzlibrary{shapes.geometric}
\usepackage{placeins}

\tikzset{
	start/.style={shape=circle, draw, minimum width=1cm, black, fill=white, thick},
	process/.style={shape=rectangle, draw, thick, minimum height=1cm, minimum width=1cm, fill=white, font=\small, text width=3cm, blue, fill=white, rotate=0},
	decision/.style={shape=rectangle, draw, minimum height=2.5cm, minimum width=2.5cm, rotate=45, font=\small,
	text width=1.75cm, fill=white ,thick},
	ecision/.style={shape=rectangle, minimum height=2.5cm, minimum width=2.5cm, rotate=0, font=\footnotesize,
	text width=1.75cm ,thick},
	line/.style={rounded corners=1cm, thick, ->, thick},
	lline/.style={white}
}

\journal{Pattern Recognition}
\begin{document}
\begin{frontmatter}




\title{Structuring the Processing Frameworks for Data Stream Evaluation and Application}


\author{Joanna Komorniczak\corref{}}
\ead{joanna.komorniczak@pwr.edu.pl}
\cortext[]{Corresponding author}
\author{Paweł Ksieniewicz}
\ead{pawel.ksieniewicz@pwr.edu.pl}
\author{Paweł Zyblewski}
\ead{pawel.zyblewski@pwr.edu.pl}

\affiliation{
    organization={Department of~Systems and Computer Networks, Wroclaw University of~Science and Technology},
    addressline={wyb. Wyspianskiego 27},
    city={Wroclaw},
    postcode={50-370}, 
    country={Poland}
}


\begin{abstract}
The following work addresses the problem of~\emph{frameworks} for data stream processing that can be used to~evaluate the solutions in~an environment that resembles real-world applications. The definition of~structured frameworks stems from a~need to~reliably evaluate the data stream classification methods, considering the constraints of~\emph{delayed} and \emph{limited} label access. The current experimental evaluation often boundlessly exploits the assumption of~their complete and immediate access to~monitor the recognition quality and to~adapt the methods to~the changing concepts. The problem is leveraged by reviewing currently described methods and techniques for \emph{data stream processing} and verifying their outcomes in~\emph{simulated environment}. The effect of~the work is a~proposed taxonomy of~\emph{data stream processing frameworks}, showing the linkage between \emph{drift detection} and \emph{classification} methods considering a~natural phenomenon of~\emph{label delay}.
\end{abstract}




\end{frontmatter}

\section{Introduction}
\label{sec:intro}

From the \emph{applied artificial intelligence} perspective, the most important factor of~data is its \emph{representation}. Even the shortest peek into the reality shows us that the dominating modality of~the contemporary internet is a~\emph{media stream} -- either in~the form of~\emph{YouTube} clips, \emph{Netflix} movies, or~\emph{TikTok} and \emph{Instagram} infinite information cascades~\cite{holyst2024protect} -- since out of~5.45 billion of~worldwide internet users, nearly 95\% consume social media~\cite{petrosyan2023worldwide}. Each of~those media streams can be, and \emph{de facto} is, processed~\cite{pigni2016digital} by the \emph{machine learning} environment as a~\emph{data stream}~\cite{ramzan2023comprehensive}.

\paragraph{Data streams and concept drift}

\emph{Data stream} is defined as an ordered collection of~objects flowing into the system in~time. The objects may be observed separately -- in~online processing -- or~aggregated into groups -- in~batch processing \cite{krawczyk2017ensemble}. A typical phenomenon visible in~the potentially infinitely incoming data stream is the time-varying \emph{probability distribution} of~represented problem \cite{kuncheva2004classifier}. It~indicates that any classification model used to~\emph{generalize knowledge} and \emph{infer decisions} in~the streaming environment depends on~the knowledge acquired so far, and on~changes yet to~come. Such changes are generally categorized as \emph{concept drifts}, most often divided into the \emph{real} and \emph{virtual} ones~\cite{benni2024impact}.

\begin{figure}[!htb]
    \centering
    \includegraphics[width=0.9\columnwidth]{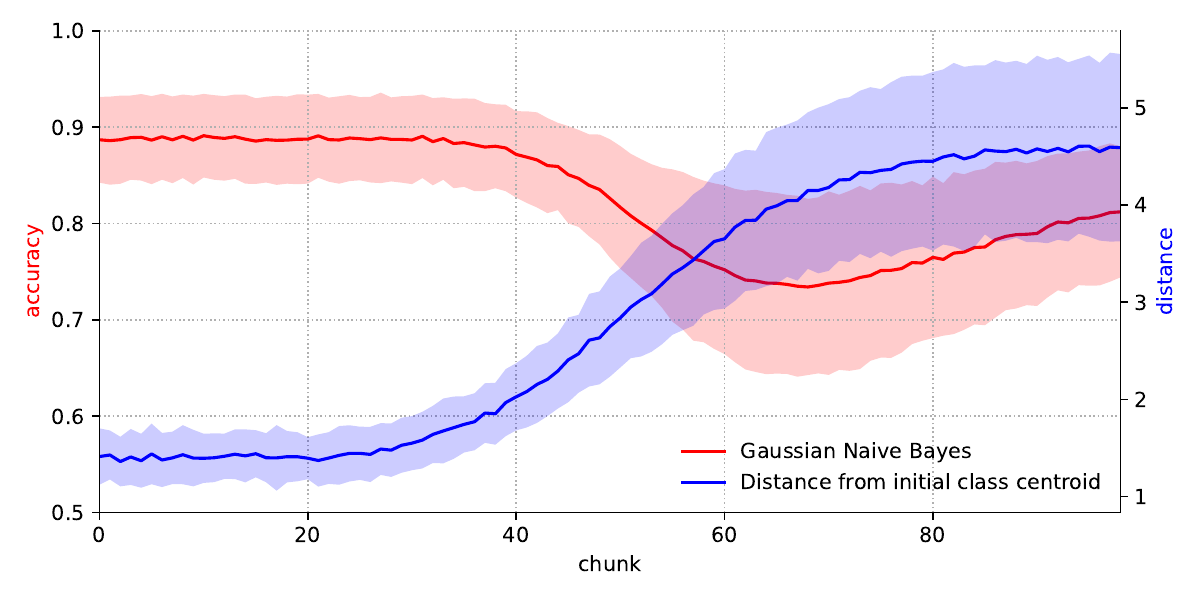}
    \caption{Averaged classification accuracy over 100 replications of~a synthetic data stream with the real drift -- in~which the initial 30 chunks do not yet show the loss of~recognition quality, despite the proceeding concept transition indicated by a~shift in~class centroids}
    \label{fig:virtual-pre}
\end{figure}

\emph{Real concept} drifts are easy to~define by directly impacting the measured quality of~a model, whereas \emph{virtual drifts} can be hindered and possibly never lead to~any degeneration of~predictive system~\cite{li2024concept}. Some guidelines for the design of~such systems even recommend relying solely on~the impact of~real drifts, ignoring the possible \emph{delayed cost factor} induced by detectable changes of~an initially virtual nature that eventually reach the status of~real drifts~\cite{cacciarelli2024active}. Figure \ref{fig:virtual-pre} shows an average recognition accuracy of~a data stream with synthetic gradual drift \cite{ksieniewicz2022stream}, in~which the first phase of~a concept change does not yet impact the classification accuracy. Meanwhile, the class distribution shift starts prior to~a concept change. The presented example shows that considering both the real and virtual concept drift is significant for the overall system recognition quality.

The minor axes of~concept drift taxonomy lean on~drift dynamic (static, gradual, and incremental drift)~\cite{ksieniewicz2022stream}, the prior class probability changes \cite{komorniczak2021prior}, or~the concept recurrence \cite{gunasekararecurrent}.

\paragraph{Classification of~data streams}

An incremental adaptation to~changes in~non-stationary data streams is a~necessary element of~effective data stream processing \cite{domingos2003}. \emph{State-of-the-art} solutions for such a~problem often follow the paradigms of~\emph{classifier ensembles} \cite{krawczyk2017ensemble}.  In~such a~system, instead of~one intelligent agent, the task is performed by the \emph{ensemble} of~independent models, whose decisions are integrated during \emph{fusion} stage \cite{wozniak2013hybrid}.

There are various strategies that enable incremental learning in~ensemble approaches dedicated to~data stream processing. One can mention the canonical method of~\textit{Streaming Ensemble Algorithm} (\textsc{sea}) \cite{street2001streaming}, in~which classifiers are trained on~separate data batches and added to~the ensemble, where they are then no longer updated. Another approach was used in~the \emph{Dynamic Weighted Majority} (\textsc{dwm}) algorithm \cite{kolter2007dynamic}, which follows incremental learning of~ensemble members. Depending on~the learning strategy used by the ensemble method, the techniques will be able to~adapt to~changing environments at~different rates. 

Currently, the most competent data stream classification methods use in-build algorithms as the \emph{base classifiers} of~homogeneous ensembles organized in~parallel architecture with a~dynamic \textit{fuser} and pruning criterion connected with a~drift detector. Classifier ensembles offer an accurate path for stably improving the general predictive power of~the in-build methods, such as \emph{Concept-adapting Hoeffding Tree}~\cite{hulten2001mining} or~\emph{Multilayer Perceptron}~\cite{rumelhart1986learning}.

As identified in~the literature, there are \emph{hybrid} processing methods that do not rely on~any internal drift detection module -- their processing strategy is denoted as \emph{passive} adaptation or~\emph{continuous rebuild} \cite{wang2003mining,brzezinski2011accuracy,wozniak2023active}. On the contrary, \emph{active} adaptation or~\emph{triggered rebuild} is visible in~methods that compose a~unit dedicated to~the detection of~a concept drift \cite{bifet2010leveraging,cano2022rose}. A drift detector is an important agent in~the architecture of~classifier ensembles of~this type. If its internal state exceeds the activation threshold, the whole ensemble does experience the occurrence of~a \emph{concept drift} alert~\cite{bifet2009improving}.

\paragraph{Drift detection and delayed labeling}

The drift detection module plays a~vital role in~\emph{hybrid} methods with active adaptation. However, over the years, a~vast pool of~solutions stating an independent tool was proposed, dedicated directly to~detecting changes in~the data distribution \cite{agrahari2022concept}. Such solutions -- denoted as \emph{concept drift detectors} -- monitor different data characteristics to~provide a~binary decision if the concept drift occurred or~if the concept remained stable.

\emph{Drift detectors} follow two canonical taxonomy branches, following the systematical axis of~\emph{real} and \emph{virtual drifts}~\cite{lobo2020spiking} as \emph{explicit} or~\emph{implicit} methods~\cite{gozuaccik2021concept}. The \textit{explicit} drift detectors will rely directly on~the quality of~the monitored recognition model, while \textit{implicit} ones will use other data characteristics.  In~terms of~\emph{label access}, all \textit{explicit} methods will be \emph{supervised}, while some of~another category -- \emph{unsupervised}. It~is worth noting that there exist \emph{implicit} detectors that require labels in~order to~monitor conditional data distribution \cite{komorniczak2023complexity}.

 In~the data streams, where the data distribution is time-dependent, we will use an indice $i$ to~denote a~position of~sequences of~features $X$ and labels $y$, describing the samples incoming in~the continuous data stream. Most often, concept drift detection methods utilize a~processing technique identified as \emph{moving window mechanism} -- objects are analyzed for some time, depending on~the structure of~a given solution. Such a~processing technique is a~cornerstone of~modern methods -- to~take as an example the \emph{Adaptive Windowing} (\textsc{adwin}) method~\cite{bifet2007learning}, used for years as a~default \emph{drift detector} in~newly designed ensemble data stream classifiers, such as \textit{Leveraging Bagging} \cite{bifet2010leveraging} and \textit{Robust Online Self-adjusting Ensemble} \cite{cano2022rose}. 

Most often, the proposed architectures of~methods assume the instant presence of~a label and, consequently, the instant possibility of~establishing the current predictive power of~a model.  In~practical applications, a~label is assigned to~the sample after some time \cite{plasse2016handling}. This property of~a \emph{data stream}, often overlooked due to~limited computer simulation resources, is denoted as a~\emph{label delay} \cite{grzenda2020}. It~states that the description of~data -- that is, labels, but also the indicator of~the system's recognition quality -- is not guaranteed to~be present at~the moment of~initial sample occurrence $i$ \cite{gomes2017adaptive}.  In~practical applications, the value $y_i$~may, and probably will, be provided after $X_i$~when the time delay  $\delta$~has passed.

As this work focuses on~processing protocols in~terms of~label requirements, the existing drift detection methods will be categorized into three groups:
\begin{itemize}
  \item \emph{Supervised} drift detector $DD_{s}$ will require class labels to~be present. It~detects a~concept change by processing a~data batch or~sample $X_i$~and a~corresponding set of~labels $y_i$. 
    \begin{equation}
        DD_{s} = f(X_i, y_i)
    \end{equation}
     In~such a~case, $X_i$~may activate the supervised drift detector only as soon as its label is observed by the system after the delay of~a label presence. The statistic on~which the decision of~most \emph{state-of-the-art} supervised detectors like \emph{Drift Detection Method} (\textsc{ddm}), \emph{Early Drift Detection Method} (\textsc{eddm}) or~\textit{Adaptive Windowing} (\textsc{adwin}) is purely related to~the recognition quality. Such explicit methods will require labels $y_i$~to~compare them with predictions $y'_i$ returned by the classifier for data $X_i$. 

    \item \emph{Unsupervised} drift detector $DD_{u}$ will only require the presence of~data $X_i$~to~provide information about the concept drift occurrence. 
    \begin{equation}
        DD_{u} = f(X_i)
    \end{equation}
    The particular benefit resulting from using an unsupervised drift detection method is the independence of~the label access. 

The cost and time delay of~labels is currently viewed as a~significant limitation of~data stream processing methods \cite{sethi2015don}, where both the velocity and the volume of~data are significant \cite{gaber2005mining}. Among the unsupervised drift detection approaches, one can distinguish the techniques that monitor the data distribution using statistical tests \cite{sobolewski2013comparable}, using one-class classifiers \cite{gozuaccik2021concept} or~that monitor the shifts of~class centroids \cite{klikowski2022concept}.

    \item \emph{Partially unsupervised} drift detector $DD_{p}$ is a~particular case of~unsupervised drift detector that requests the labels $y$ exclusively in~the event of~a concept change. The method operates similarly to~$DD_{u}$ at~a~regular mode by only processing data features $X_i$. However, it is possible that the detector requires labels $y_i$, which makes it operate similar to~$DD_{s}$ in~certain states of~a system -- directly after the drift detection. Therefore, depending on~the current state, the detection can be performed in~two modes: supervised or~unsupervised.
    
    \begin{equation}
        DD_{p} = f(X_i, y_i)
    \end{equation}
    
    \begin{equation}
        DD_{p} = f(X_i)
    \end{equation}

    This category is composed of~methods that require to~be \emph{re-calibrated} after the drift occurrence. Compared with $DD_{s}$, the label request may be sent sporadically only when the concept drift is detected and the model rebuild is required. The methods that monitor the properties of~the underlying classification model -- other than the classification quality reserved for explicit and supervised solutions -- include many approaches from this group. \textit{State-of-the-art} drift detectors employ monitoring the density of~samples near the decision boundary \cite{sethi2015don}, the support function of~classification model \cite{lindstrom2013drift}, or~the explainability of~its decisions \cite{haug2022change}.
    
\end{itemize}

\paragraph{Contribution and Motivation}

This work looks at~the current methods and processing protocols from a~\emph{big picture perspective}, considering the real-world mechanisms that need to~be employed for incremental learning and the implications of~such mechanisms on~the classification quality of~\textit{state-of-the-art} methods. 

The main contributions of~the presented work are:
\begin{itemize}
    \item The presentation of~four structured \emph{frameworks} presenting the paradigms of~data stream processing when using different categories of~drift detection methods. 
    \item Consideration of~unavoidable \emph{label delay} phenomenon in~the presented processing protocols.
    \item The experimental evaluation of~\textit{state-of-the-art} in-build methods coupled with different types of~drift detectors, showing the implications of~different processing schemes.
    \item Comparison of~existing methods in~a~context of~various label delay times and concept drift frequencies.
    \item Consideration of~an abstract \emph{Oracle} drift detector in~the experiments, limiting the uncertainty related to~the correct drift detection by existing methods.
    \item Extending the standard evaluation of~(a) classification quality while additionally taking into account (b) the number of~label requests and (c) the frequency of~classifier training.
\end{itemize}

The proposed contributions will support the proper and fair experimental evaluation of~methods, allowing for considering the impact of~label cost, their delay, and the cost of~model rebuild in~the event of~a concept change. Those matters are often neglected in~comparing methods presented in~the research but remain significant in~real-world applications \cite{grzenda2020,costa2018drift,bartz2024drift}.

\section{Frameworks}
\label{sec:frameworks}

The presented frameworks aim to~systematize the data stream processing schemes used in~the application and evaluation of~methods. Those frameworks consider the context of~the drift detection unit and its type in~terms of~label requirements. Each of~the presented schemes considers a~\emph{label delay} factor.

Depending on~the type of~drift detector used, the conceptual processing scheme will vary, especially in~the context of~accessing labels and updating the classifier. Such processing frameworks may be generalized into the four categories: \textit{(a)} \emph{continuous rebuild} -- using passive adaptation to~current data, not relying on~drift detection module, and \textit{(b)} \emph{triggered rebuild} -- actively adapting to~the current concept in~the presence of~noticed factors that may indicate a~change in~the distribution of~processed data. The second can be categorized further into those using  \emph{supervised}, \emph{unsupervised}, and \emph{partially unsupervised} drift detectors.

These frameworks will contain some common processing blocks, each having a~dedicated role in~a~processing system. The blocks identified with {\color{blue}blue} color indicate an independent process or~mechanism, while the blocks identified with {\color{red}red} color -- the conditional statements. The system's state after such a~processing block will depend on~the output of~the evaluated condition, e.g. concept drift detection.

The following processing blocks were identified:
\begin{itemize}
    \item {\color{blue}\emph{Label request} block} -- To resemble real-world applications, the presented frameworks rely on~label availability on~request. The request will be fulfilled after the time delay specified by the  $\delta$~parameter. The \emph{continuous rebuilt} and \emph{triggered rebuild} with a~\emph{supervised} drift detector will automatically request labels, while the frameworks using \emph{unsupervised} and \emph{partially unsupervised} drift detectors will request labels only in~specific states of~the processing system.
    
    \item {\color{blue}\emph{Classifier fitting} block} -- The block is dedicated to~the incremental training of~the classifier. The guidelines of~data stream processing solutions force the adaptation of~the classification method. This block is obligatory for most iterations of~the frameworks following \emph{continuous rebuild} scenario.  In~\emph{triggered rebuild}, it will be performed exclusively after the concept change is detected. It~is essential to~note that in~real-world applications, the rebuilding of~training of~a classifier demands specific computational and time resources, in~addition to~the cost of~label acquisition.
    
    \item {\color{blue}\emph{Return predictions} block} -- The block is often performed as the final step of~data sample or~data chunk processing. The current classification model returns predictions $y'_i$ for incoming data $X_i$~in~this processing block. It~is necessary for a~model to~provide an output regardless of~the current state of~the framework.
    
    \item {\color{blue}\emph{Store predictions} block} -- This block will be used only in~the case of~\emph{supervised} and explicit drift detectors, which rely on~the assessment of~the classifier's quality to~detect a~concept change. The predictions $y'_i$ will be used by a~supervised detector $DD_s$ to~compare with true labels $y_i$~provided to~the system.
    
    \item {\color{red}\emph{Check label availability} block} -- This block is dedicated to~reviewing if the labels requested in~the past have arrived.  In~the presented work, the delay time is indicated by  $\delta$~and remains constant. However, the  $\delta$~can be sampled from any distribution, describing the \emph{stochastic} or~\emph{semi-stochastic} delay \cite{plasse2016handling}. The block will indicate a~positive result (\emph{true}) when the  $\delta$~chunks have passed since the label request for the chunk $i$.
    
    \item {\color{red}\emph{Concept Drift Detection} block} -- This block is used in~\emph{triggered rebuild} processing schemes. It~uses various concept drift detection methods and, depending on~their type, utilizes the data $X_i$, extended with the labels $y_i$, and, only in~the case of~explicit drift detectors, the predictions of~the classifier $y'_i$. The block will indicate a~positive result if the drift is detected.
    
    \item {\color{red}\emph{Unfulfilled label request} block} -- This is a~unique block used when processing data with a~\emph{partially unsupervised} drift detector. The pending label request will be indicative of~a state in~which the drift detector $DD_p$ indicated a~drift by processing only the data $X_i$, but to~continue the proper detection, it needs to~be rebuilt with the use of~requested labels $y_i$. The unfulfilled label request returns a~positive outcome when the concept change has been recognized, but the drift detector is not yet ready to~properly evaluate the data in~search of~the following concept changes.
\end{itemize}

All described blocks are arranged into processing schemes, presenting the operation of~four processing frameworks for data streams with non-stationary concepts. The following paragraphs will contain a~specific description of~established mechanisms that take into account the \emph{delayed labeling} phenomena.

\subsection{\textbf{C}ontinuous \textbf{R}ebuild (CR)}

\begin{figure}[!h]
	\centering
	\resizebox{\columnwidth}{!}{
		\input{frameworks/cr}}
	\caption{The processing scheme of~Continuous Rebuild framework}
	\label{fig:cr2}
\end{figure}
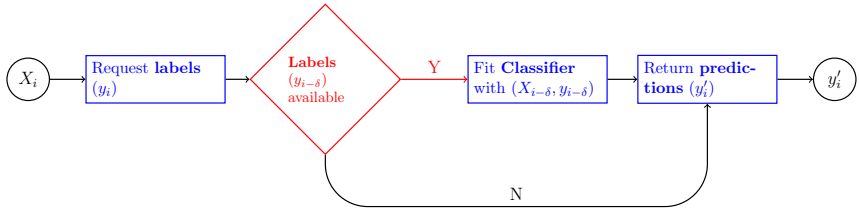

The first, most straightforward approach for evaluation follows a~\emph{continuous rebuild} paradigm. It~is based on~a~foundation of~static \emph{incremental learning}, where the underlying classifier is incrementally updated regardless of~changes in~the concept. The overall processing of~\emph{CR} is visualized in~Figure~\ref{fig:cr2}.

The first processing element is requesting labels  $y_i$~for the current chunk $X_i$. Those will be necessary for the classifier training. The next step is to~check whether previously requested labels for data from the chunk $X_{i-\delta}$ have been delivered. If $y_{i-\delta}$ labels are available, a~training set is created from the features $X_{i-\delta}$, described by $y_{i-\delta}$ labels. The final step of~the processing framework is to~return the predictions for $X_i$.

\subsection{\textbf{T}riggered \textbf{R}ebuild with \textbf{S}upervised Drift Detection (TR-S)}

\begin{figure}[!h]
	\centering
		\resizebox{\columnwidth}{!}{
		\input{frameworks/tr-s}
	}
	\caption{The processing scheme of~Triggered Rebuild with Supervised Drift Detection framework}
	\label{fig:tr-sup2}
\end{figure}
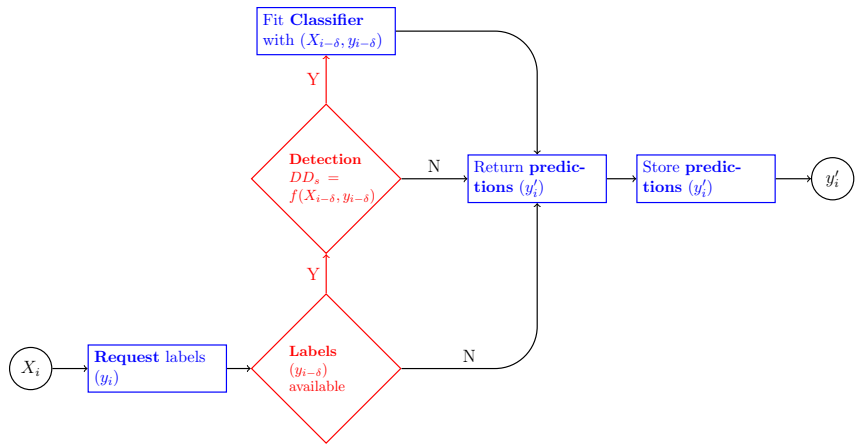

A more complex approach for data stream processing follows the \emph{triggered rebuild} strategy, in~which the \emph{supervised} drift detection method is used to~indicate the visible concept changes in~the data. As in~all frameworks following the \textit{triggered rebuild} strategy, the classifier training will be performed only in~the case of~detected concept change. The overall processing scheme using \emph{TR-S} is visualized in~Figure~\ref{fig:tr-sup2}.

 In~this processing scheme -- similarly to~\emph{Continuous Rebuild} -- the labels $y_i$~are requested for each data chunk $X_i$. However, in~the previous framework, they were used only for incremental training, whereas here, they are necessary for supervised concept drift detection. The next step is to~check whether the labels requested in~the chunk $i-\delta$ have arrived -- that is if the time  $\delta$~since their request in~chunk $i$ has passed. If so -- a~set of~features, labels, and -- optionally for explicit detectors -- past predictions are prepared. Based on~those values, the drift detection using $DD_s$ is performed. 

As the detector evaluates data describing chunk $i-\delta$, the provided state of~the detector will represent the moment from the past. If the change is detected, the base classifier is trained with the past data and past labels from the chunk $i-\delta$. Despite the action being performed in~the chunk $i$, the samples used to~fit the model describe the chunk from time point $i-\delta$. The last step, common regardless of~whether the detector has signaled a~drift, will be to~return the predictions $y'_i$ for the current chunk. 

\subsection{\textbf{T}riggered \textbf{R}ebuild with \textbf{U}nsupervised Drift Detection (TR-U)}

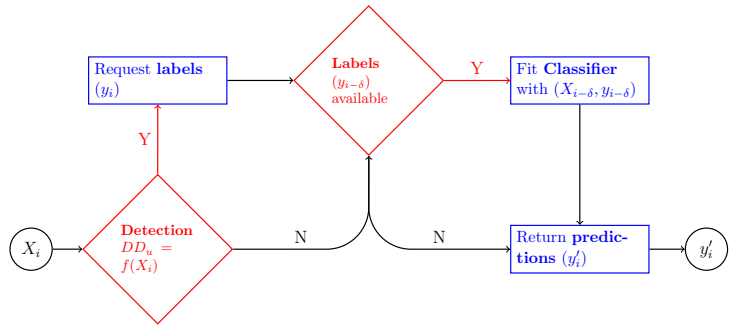
\begin{figure}[!h]
	\centering
	\resizebox{\columnwidth}{!}{
		\input{frameworks/tr-u.tex}
	}
	\caption{Processing diagram of~Triggered Rebuild with Unsupervised Drift Detection framework}
	\label{fig:tr-uns2}
\end{figure}

Another variation of~\emph{triggered rebuild} is using the \emph{unsupervised} drift detector. Compared with the previous approach (TR-S), the concept change can be recognized regardless of~label delay $\delta$, as the drift detection method $DD_u$ only analyses the features of~data $X_i$. The labels, however, will still be required for the classifier update, and, as in~any real-world application, the time delay  $\delta$~will impact such a~process. The overall processing of~\emph{TR-S} is visualized in~Figure~\ref{fig:tr-uns2}.

The first step of~such a~processing scheme is checking if the concept drift occurred using the dedicated $DD_u$. As mentioned, the great advantage of~this processing scheme is the lack of~requirement for labels during detection, which means that changes are detected immediately, and the detection step can be performed for every chunk. If drift is detected, a~request for labels $y_i$~is made. The following step is to~check if the samples requested in~the past chunk $i-\delta$ are available. If the required time  $\delta$~has passed, the classifier is fitted with past data $X_{i-\delta}$ and obtained labels $y_{i-\delta}$.

\subsection{\textbf{T}riggered \textbf{R}ebuild with \textbf{P}artially Unsupervised Drift Detection (TR-P)}
 
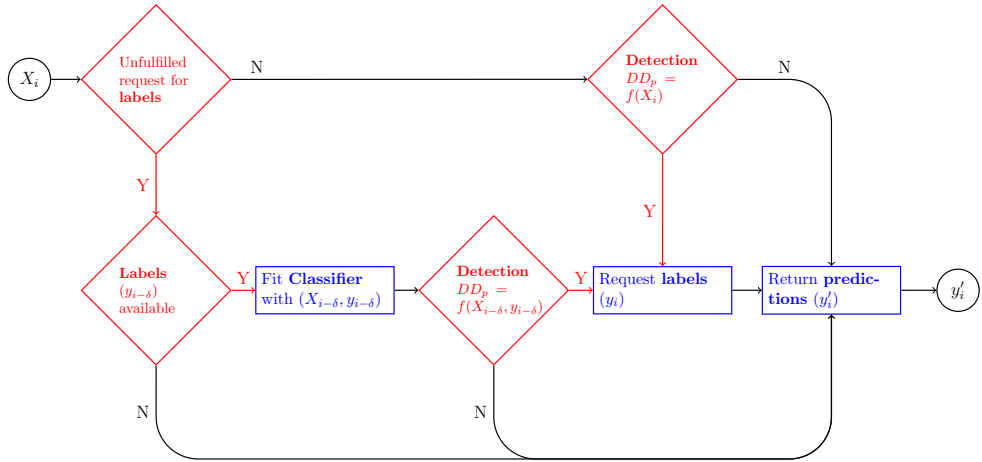
\begin{figure}[!h]
	\centering
	\resizebox{\columnwidth}{!}{
		\input{frameworks/tr-p}
	}
	\caption{Processing diagram of~Triggered Rebuild with Partially Unsupervised Drift Detection framework}
	\label{fig:tr-sem2}
\end{figure}

This framework uses \emph{partially unsupervised} drift detector, which, despite the unsupervised detection, will still require labels after a~concept change to~enable the detection of~the subsequent concept changes. This can be understood as \emph{fitting} the detector to~the new concept, which, similarly to~a~classifier, is required for the ability of~a method to~recognize a~drift. The overall processing of~\emph{TR-P} is visualized in~Figure~\ref{fig:tr-sem2}.

 In~this processing scheme, the first block checks if there is an unfulfilled label request, possibly resulting from a~past detection. Such a~check is necessary to~determine if the detector $DD_p$ is competent to~detect changes in~unsupervised mode $DD_p = f(X_i)$ or~needs to~be fitted to~the new concept by additionally utilizing the labels $DD_p = f(X_i, y_i)$.

The detector can operate in~an unsupervised mode when there is no past label demand. If a~drift is detected, similarly to~other processing schemes, a~request for labels is made. Later, the base classifier returns predictions for the current chunk $X_i$.

 In~case of~a pending label request, the next block checks if labels requested in~the past $y_{i-\delta}$ are available. If so, the classifier is first trained using the past data $X_{i-\delta}$ and the received labels $y_{i-\delta}$. Next, the detector receives labels, allowing for its future fully unsupervised detection. It~is highly improbable that the detector will indicate concept drift at~this point, as the purpose of~providing the labels is to~re-adjust the classifier to~the new concept. However, to~follow the processing paradigm, we assume that drift can be indicated by $DD_p$ regardless if it used only features $X_i$, or~both features and labels $(X_{i-\delta}, y_{i-\delta})$. It~is important to~note that the labels will arrive for the chunk $i-\delta$, hence, the $DD_p$ using the labels will analyze the past data.  In~contrast, the detection based solely on~features can be made for the current data chunk $i$. As in~all previously described frameworks, the final processing step is to~return the predictions $y'_i$ for the current data. 

\vspace{0.5em}
By establishing such processing schemes using common \emph{blocks}, we aim to~underline the importance of~their usage in~computer experiments, simulating the real-world systems for which the data stream processing methods are designed. 

\section{Experimental set-up}
\label{sec:exp}

The following section contains an experiment showing how \textit{state-of-the-art} classification methods perform in~four environments described in~Section \ref{sec:frameworks}, depending on~(\emph{a}) the frequency of~concept drifts in~the stream and (\emph{b}) various delay  $\delta$~in~the arrival of~labels. The experimental code, the implementation of~frameworks and the configuration of~the environment are publicly available in~a~\textit{GitHub} repository\footnote{\url{https://github.com/w4k2/skipper}}.

It is easy to~state that the number of~concept changes in~a~given stream processing period (i.e., the drift frequency) and the delay time  $\delta$~will impact the classification quality. Additionally, the correlation between the frequency of~concept drift and the delay  $\delta$~is expected to~be visible. This will happen when the time between the concept changes is shorter than the delay of~labels.  In~such a~case, the classifier, despite the rebuild process, will not be adjusted to~the currently processed concept -- but the previous one, from which the labels arrived.

\subsection{Data streams}

Data streams used in~the experiment were synthesized using the generator from the \textit{stream-learn} library \cite{ksieniewicz2022stream}, which allowed for precise configuration of~the frequency of~concept drifts and their dynamics. An accurate ground truth of~the drift occurrence moments is only available in~the case of~synthetic streams, which is an essential advantage of~using this type of~data~\cite{bifet2015efficient}.

The performance of~the \textit{state-of-the-art} methods was evaluated for the case of~$5$, $10$, and $15$ sudden drifts occurring over $500$ chunks of~the data stream.  In~the data stream, each chunk contains information about $250$ objects described by $20$ quantitative features, $15$ of~which are informative. One of~the significant advantages of~evaluating methods on~synthetic data is the possibility of~multiple replications of~data generation, which makes it possible to~stabilize the results obtained using the tested methods~\cite{stapor2021design}.  In~the conducted research, each stream was generated ten times and evaluated multiple times in~each specified environment.

\subsection{Delay factor}

The evaluation was performed individually for four described frameworks: \textit{Continuous Rebuild} (CR) and \textit{Triggered Rebuild with Supervised} (TR-S), \textit{Unsupervised} (TR-U), and \textit{Partially Unsupervised} (TR-P) drift detectors. Each processing environment was evaluated for four different delay times $\delta$.  In~the presented research, the delay time was constant and took values of~$1$, $10$, $20$, and $60$ chunks. The constant delay times were chosen as the simplest and most illustrative scenario that limits the uncertainty factor of~the results. However, the presented frameworks are well suited for conducting research with \emph{stochastic} or~\emph{semi-stochastic} $\delta$, sampled from a~distribution of~researcher's choice \cite{plasse2016handling}.

For a~delay time equal to~one, the frameworks will work analogously to~the commonly used \emph{test-then-train} protocol \cite{ksieniewicz2022stream}, which assumes an immediate inflow of~labels delivered right after the classification quality testing procedure. The highest value of~the  $\delta$~parameter will show how the methods behave when the interval between the concept changes is smaller than the delay time of~label arrival -- such a~phenomenon will be visible for streams with $10$ and $15$ drifts.

\subsection{Framework components}

The presented frameworks describe only the protocols of~data processing. It~means that any of~them needs to~be parametrized with a~classification method capable of~incremental learning and -- in~the case of~\emph{triggered rebuild} frameworks -- the concept drift detection method that follows a~specific processing paradigm. 

 In~order to~limit the bias related to~the quality of~detection in~the presented experiments, two solutions were evaluated: 
\begin{enumerate}
    \item[(\emph{a})] using \emph{state-of-the-art} detection methods from each branch of~taxonomy, respectively:
	\begin{itemize}
		\item \emph{Drift Detection Method}~\cite{gama2004learning} (\textsc{ddm}) as a~supervised detector, 
		\item \emph{One-class Drift Detection}~\cite{gozuaccik2021concept} (\textsc{ocdd}) as an unsupervised detector, 
		\item \emph{Margin Density Drift Detection}~\cite{sethi2015don}(\textsc{md3}) as a~partially unsupervised detector.
	\end{itemize}
 
	\item[(\emph{b})] using \emph{Oracle} drift detector, which is an abstract model that will always correctly determine the moment of~drift. The inspiration for this abstract was taken from the research in~the branch of~classifier ensembles \cite{woods1997combination}, where \emph{Oracle} ensures a~correct decision by one of~the ensemble's members.  In~the context of~presented frameworks, this approach will allow for establishing results in~the hypothetical scenario where the drift detection components are never wrong. It~is worth noting that the \emph{Oracle} can be simulated only due to~the use of~synthetic data streams, which deliver a~concept drift ground truth.
\end{enumerate}

The second component necessary for each of~the frameworks is the classification method. Due to~the large number of~possible method configurations and different strategies of~both \emph{base classifiers} and \emph{hybrid methods} for continuous adaptation, it was decided to~compare three autonomous in-build models capable of~incremental learning, classically used as components of~ensemble methods. Such an investigation of~the ability of~base learners to~adapt to~concepts at~different label delay times will allow for an assessment of~the overall ability to~process non-stationary data streams, which should translate into such ability in~individual ensemble methods, desensitizing the experiment to~the specific and detailed mechanisms used in~those approaches.

The following three classification methods were evaluated:
\begin{itemize}
      \item \textit{Gaussian Naive Bayes} (\textsc{gnb}) \cite{anderson1992explorations} -- a~method that has a~natural ability to~process samples incrementally. The classifier updates the class-conditional probabilities throughout the processing  \cite{kuncheva2004classifier}. As all data in~incremental learning have equal weight, the method will not be well adapted to~the specific concept -- rather, it will induce knowledge based on~all previously learned concepts.

     \item \textit{Multilayer Perceptron} (\textsc{mlp}) \cite{hinton1990connectionist} -- a~method capable of~incrementally updating the weights of~the underlying model. Due to~the \emph{catastrophic interference} \cite{french1999catastrophic}, a~newly learned concept will be of~more significance in~the current decision than past concepts (unlike \textsc{gnb}).  In~the case of~the CR approach, the method was trained in~a~single iteration.  In~the case of~TR -- when training occurs less frequently and only as a~result of~signaling a~change in~a~concept -- the classifier was trained in~50 iterations.
     
     \item \textit{Hoeffding Tree Classifier} (\textsc{ht})~\cite{hulten2001mining} -- a~method also denoted as \emph{Concept-adapting Very Fast Decision Trees}, which offers an adaptation of~the Decision Tree training process for the incremental scenario of~data stream processing and the possibility of~adaptation of~the tree structure. The method uses Hoeffding inequality to~calculate the sample size needed to~determine the splitting criterion, which allows building the tree structure with a~single pass of~the available data.
     
\end{itemize}

The implementation of~the classification methods was taken from the \emph{scikit-learn}~\cite{scikit-learn} and \emph{scikit-multiflow}~\cite{skmultiflow} libraries, using their default hyperparameters.

\subsection{Evaluated criteria}

 In~the experiments, we measured the most common evaluation criteria -- the classification quality. The chosen metric was \emph{balanced accuracy} (\textsc{bac}). Additionally, the experiments considered the number of~label requests reported by each framework in~evaluated data stream scenarios and the number of~classifier rebuild demands. Those two additional criteria were motivated by the implications of~using the presented frameworks in~a~real-world setting -- where both the cost of~sample annotation and the cost of~model training may be significant. Therefore, the three criteria used to~evaluate the processing scenarios were:
\begin{itemize}
    \item \textit{balanced accuracy} metric -- this measure for the evaluated type of~balanced streams will work analogously to~accuracy and describe the classification quality in~each chunk. This measure is expected to~be especially impacted by the  $\delta$~of~delayed labeling.
    
    \item the proportion of~iterations where \textit{label request} was signaled -- this measure will describe the impact of~label cost on~the processing. Most evaluation protocols assume complete access to~labels. Such a~non-realistic presumption would mean that each processed sample will receive the correct description at~some point, which is highly improbable in~real-world applications. The frameworks that follow a~CR processing strategy or~use a~\emph{supervised} drift detection module will naturally depict a~high cost of~label acquisition.
    
    \item the proportion of~iterations where \textit{training the classifier} was performed -- this measure will consider the cost of~classifier training. Similarly to~label cost, the impact of~a model rebuild is rarely considered when evaluating the data stream processing methods. Meanwhile, despite the availability of~labels, the training process could be skipped if seen as redundant or~unnecessary. Such a~scenario will be visible in~all TR processing frameworks when -- in~case the concept remains static -- there will be no need to~rebuild the classifier.
\end{itemize}

The classifier was always trained in~the first iteration (with the first data chunk) of~all evaluation runs, regardless of~the processing and drift detection scheme. Additionally, a~label request occurs in~the first processing chunk, except for the TR-P protocol, where labels are required after drift detection.

\section{Results of~experiments}
\label{sec:results}

\begin{figure}[!htb]
    \centering
  \includegraphics[width=\columnwidth,clip=false,trim=20 10 20 10]{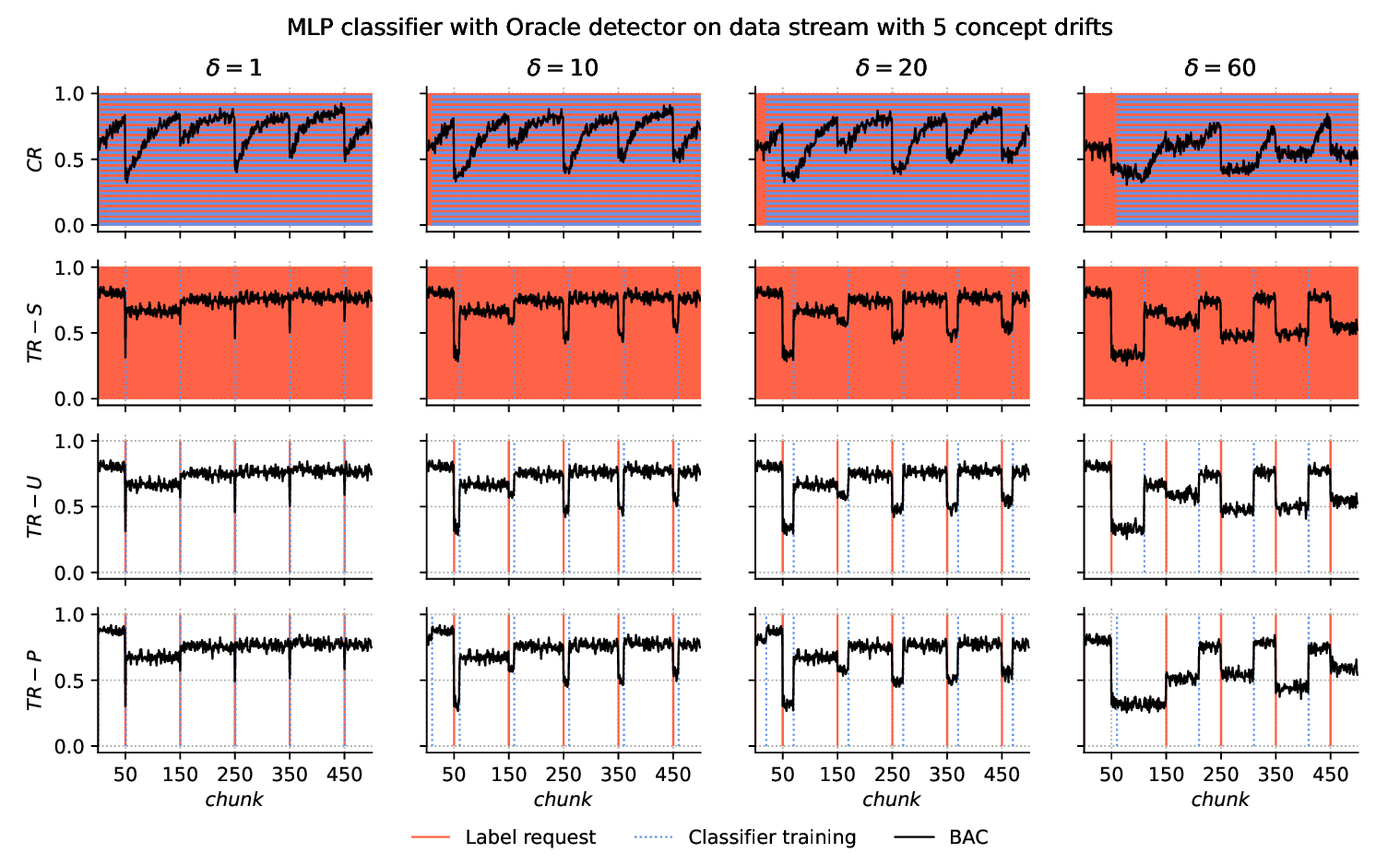}
    \caption{The balanced accuracy metric, moments of~label requests, and classifier training for a~single data stream with five concept changes. The moments of~concept changes are signaled with ticks on~the \textit{x} axis. The rows present different data stream processing frameworks, and the columns present the results for different delta parameter values. Red and blue vertical lines present label request and classifier training moments, respectively.}
    \label{fig:acc}
\end{figure}

The results from a~single stream flow on~various framework configurations are presented in~Figure~\ref{fig:acc}. The consecutive rows present four approaches to~stream processing: \textit{Continuous Rebuild} (CR), \textit{Triggered Rebuild with Supervised Drift Detection} (TR-S), \textit{Triggered Rebuild with Unsupervised Drift Detection} (TR-U), and \textit{Triggered Rebuild with Partially Unsupervised Drift Detection} (TR-P). 

The columns indicate the quality changes over different values of~the  $\delta$~parameter, describing the label delay -- from 1 to~60. Each sub-plot shows the quality of~classification measured as the balanced accuracy, the moments of~requesting labels (vertical red lines), and the moments of~rebuilding the classifier (vertical blue lines). The figure shows the results for the \textsc{mlp} classifier and the synthetic \textit{Oracle} drift detector, which always indicates the correct moments of~drift occurrence.

Observation of~the classification quality makes it trivial to~see that the label delay ($\delta$) is a~significant factor for the overall model efficiency -- even with perfect detection, the classification quality will be deficient until the labels are delivered and the classifier is rebuilt.  In~the case of~CR (first row), we see a~smooth increase in~recognition quality as the \textsc{mlp} classifier is incrementally trained until a~concept change occurs.  In~the case of~the remaining frameworks (TR), the classification quality for a~stable concept is constant, as the classifiers are not trained. 

It is worth mentioning that in~the case of~a  $\delta$~parameter value of~$60$, the classification quality after the first drift returns to~a high value only in~approximately chunk $210$. This is due to~the request for labels for the detector in~the first processing chunk. The detector waiting for labels could not recognize the first drift, and only after the second change (occurring in~the 150th chunk of~the stream) and additional $60$ chunks ($\delta$) did the classifier adapt to~the currently occurring concept.  In~the case of~streams with $10$ and $15$ drifts and the highest tested $\delta$, the concept will change more often than the label delay time, significantly affecting the recognition quality.

Another easily noticeable component of~the results in~Figure~\ref{fig:acc} is the number of~label requests and classifier rebuilds, indicated by the red and blue vertical lines.  In~the case of~\textit{continuous rebuild} and \textit{triggered rebuild}, the request occurs in~each processing chunk -- for classifier training (CR) or~drift detection (TR-S), respectively. It~is worth noting that both of~these operations will be delayed by a~ $\delta$~in~relation to~the actual concept change -- in~the case of~TR-S, the classifier is trained immediately after the change is detected. However, the labels used for detection may already be out of~date.  In~this protocol, compared to~CR, the classifier is only rebuilt when drift is recognized, indicated by single vertical blue lines.  In~the case of~TR-U and TR-P, visible in~rows 3 and 4 of~the presented figure, both the label request and the rebuilding of~the classifier take place only after the drift detection and after the delivery of~labels.

\begin{figure}
    \centering
    \includegraphics[width=\columnwidth]{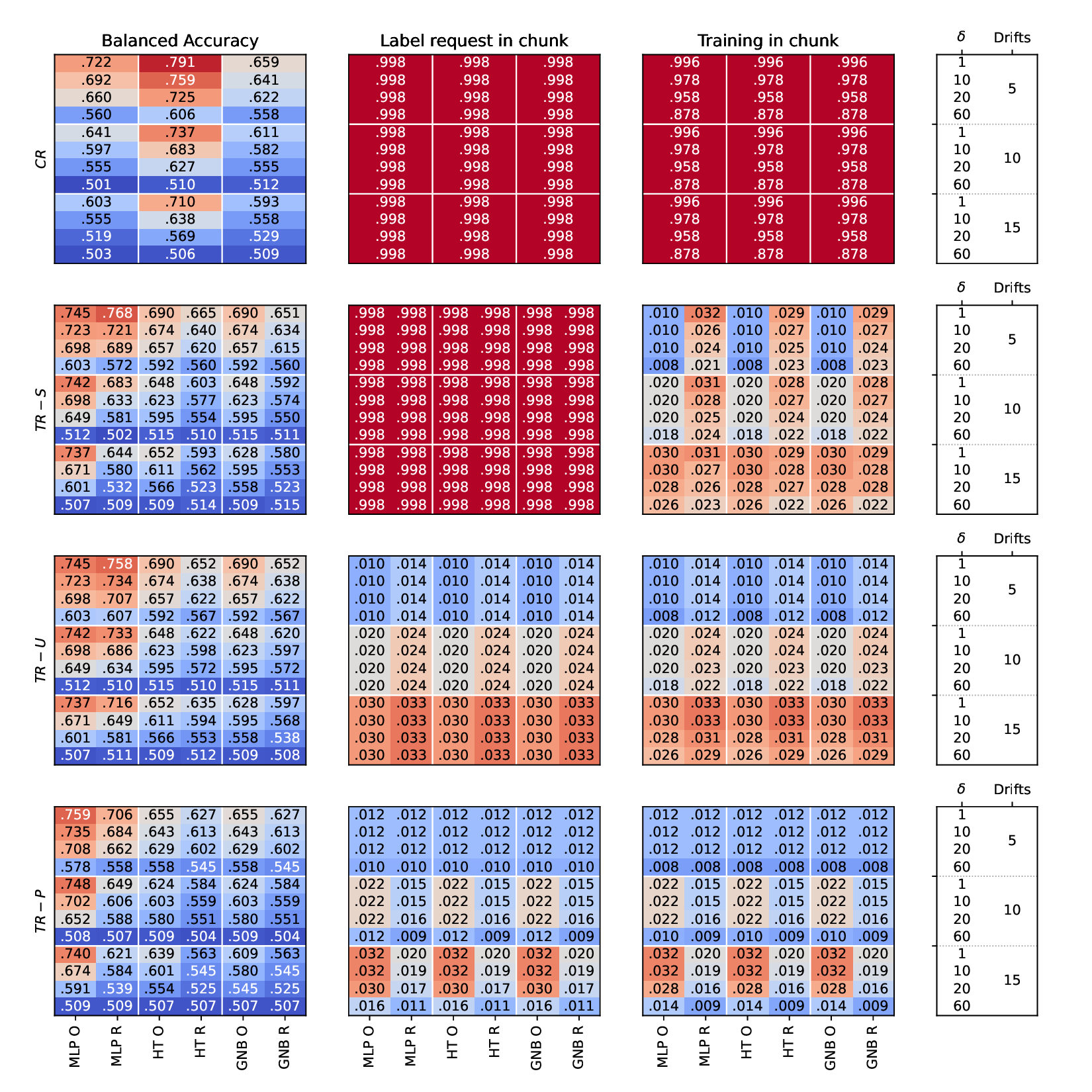}
    \caption{The color-coded results of~the experiment for all evaluated classification and drift detection methods (\textit{x} axis of~subplots) and all evaluated data stream configurations (\textit{y} axis of~subplot). The columns present an averaged balanced accuracy, the fraction of~chunks with label request, and the fraction of~chunks with classifier rebuild. The red cell is associated with a~high measure value, and blue with a~low value.}
    \label{fig:tab}
\end{figure}

The results for all data streams are presented as a~color-coded table in~Figure~\ref{fig:tab}, where the columns present the average balanced accuracy, the average percentage of~chunks with label requests, and the average percentage of~chunks with classifier rebuilding, respectively. There is a~clear advantage in~the number of~label requests and rebuilds for the CR and TR-S frameworks, which was noticed earlier. Additionally, as expected, at~$10$ and $15$ drifts and a~ $\delta$~of~$60$, the classification quality is poor -- due to~the arrival of~labels after the completion of~the concept for which the request was made. This environment configuration also influenced the number of~label requests and the number of~classifier rebuilds, resulting from fewer detections and classifier rebuilds, as some concept changes remained unnoticed while waiting for labels to~arrive. The number of~label requests and rebuilds is greatly influenced by the number of~drifts in~the stream, which results from the need to~adapt to~the subsequent concepts. 

Additionally, it can be seen that using real drift detectors, depending on~the framework, brings a~difference in~the tested values compared to~the synthetic \emph{Oracle} detector. Depending on~the detection method used, more label requests and the need to~rebuild the classifier will be signaled, e.g., in~the case of~the \textsc{ocdd} detector and the TR-U approach and the \textsc{ddm} detector for TR-S, or~less -- in~the case of~the \textsc{md3} detector and the TR-P approach. 

Using an authentic drift detector almost always results in~a~decrease in~classification quality. The exceptions to~this observation are single runs for \textsc{mlp}, where redundant detections allow training the classifier for the current concept. However, this came at~the cost of~redundant label requests and a~number of~classifier rebuilds.

\section{Conclusions}
\label{sec:conc}

This article systematized the possible schemes of~processing data streams by the \emph{in-build} and \emph{hybrid} classification methods into the solutions 
following \textit{continuous} or~\textit{triggered} rebuild approach in~the face of~concept drift.  In~the latter, all three types of~existing drift detection strategies -- \textit{supervised}, \textit{unsupervised}, and \textit{partially unsupervised} -- were considered, with the schemes reflecting their strengths and limitations. Each of~the presented frameworks aimed to~extend the processing protocols by considering costs typical to~the real-world applications of~data processing systems, such as the cost of~label acquisition and classifier training, as well as the inevitable phenomena of~\textit{label delay}.  

The proposed frameworks were used in~the computer experiments to~monitor the classification quality and the remaining evaluation criteria. For the purpose of~the experiments, we evaluated three canonical in-build classification models and three \textit{state-of-the-art} drift detectors, additionally extended with the abstract \emph{Oracle} drift detector, to~limit the bias of~drift detection quality by existing methods. Analysis shows that including a~factor of~\emph{label delay}, as well as the characteristics of~specific types of~drift detection methods clearly affects the overall efficiency of~used models, and the remaining criteria monitored in~the computer experiments.

The above work does not aim to~criticize currently established processing protocols based on~supervised drift detection, but to~extend the standard processing paradigm for additional, realistic factors. We hope that the proposed abstract processing frameworks will facilitate the design of~real-world solutions and enable the thorough evaluation of~employed methods by providing a~systematic and experimentally verified design pattern.


\section*{Acknowledgement}
This work is supported by the CEUS-UNISONO programme, which has received funding from the National Science Centre, Poland under grant agreement No. 2020/02/Y/ST6/00037 and statutory funds of the Department of Systems and Computer Networks.

\bibliographystyle{elsarticle-num} 
\bibliography{cas-refs}

\end{document}

%% file: frameworks/cr.tex
\begin{tikzpicture}


\node[start] at (0,0) (start) {$X_i$};

\node[process] at(3,0) (a) {Request \textbf{labels} ($y_i$)};
\node[decision, draw=red] at(7,0) (b) {};
\node[ecision, red] at(7,0) (bb) {\textbf{Labels} ($y_{i-\delta}$) available};
\node[process] at(12,0) (d) {Fit \textbf{Classifier} with $(X_{i-\delta}, y_{i-\delta})$};
\node[process] at(16,0) (e) {Return \textbf{predictions} ($y'_i$)};
\node[start] at (19,0) (stop) {$y'_i$};

	\draw[line] (start) -- (a);
	\draw[line] (a) -- (b);
	\draw[line] (b) -- (7,-3) --node[midway,above] {N} (16,-3) -- (e);
	\draw[line, red] (b) -- node[midway, above] {Y} (d);
	\draw[line] (d) -- (e);
	\draw[line] (e) -- (stop);


 	\draw[lline] (-2,4) -- (22,4);
	\draw[lline] (-2,-4) -- (22,-4);
	\draw[lline] (-2,4) -- (-2,-4);
	\draw[lline] (22,4) -- (22,-4);

\end{tikzpicture}

%% file: frameworks/tr-s.tex
\begin{tikzpicture}
	
    
	\node[start] at (0,0) (start) {$X_i$};

	\node[process] at(3,0) (a) {\textbf{Request} labels ($y_i$)};

	\node[decision, draw=red] at(7,0) (b) {};
	\node[ecision,red] at(7,0) (bb) {\textbf{Labels} ($y_{i-\delta}$) available};

	\node[decision, draw=red] at(7,4.5) (d) {};
	\node[ecision,red] at(7,4.5) (dd) {\textbf{Detection} $DD_s = f(X_{i-\delta}, y_{i-\delta})$};
	
	\node[process] at(7,8) (g) {Fit \textbf{Classifier} with $(X_{i-\delta},y_{i-\delta})$};

	\node[process] at(12,4.5) (e) {Return \textbf{predictions} ($y'_i$)};

	\node[process] at(16,4.5) (f) {Store \textbf{predictions} ($y'_i$)};

	\node[start] at (19,4.5) (stop) {$y'_i$};
			
	\draw[line] (start) -- (a);
	\draw[line] (a) -- (b);
	\draw[line,red] (b) --node[left,midway] {Y} (d);
	\draw[line,red] (d) --node[left,midway] {Y} (g);
	\draw[line] (g) -- (12,8) -- (e);
	\draw[line] (e) -- (f);
	\draw[line] (d) --node[above,midway] {N} (e);
	\draw[line] (b) --node[above,midway] {N} (12,0) -- (e);
	\draw[line] (f) -- (stop);


  	\draw[lline] (-2,9) -- (22,9);
	\draw[lline] (-2,-3) -- (22,-3);
	\draw[lline] (-2,9) -- (-2,-3);
	\draw[lline] (22,9) -- (22,-3);

\end{tikzpicture}

%% file: frameworks/tr-u.tex
\begin{tikzpicture}


	\node[start] at (0,0) (start) {$X_i$};

	\node[decision, draw=red] at(3,0) (a) {};
	\node[ecision,red] at(3,0) (aa) {\textbf{Detection} $DD_u = f(X_i)$};
	\node[process] at(3,4) (b) {Request \textbf{labels} $(y_i)$};
	\node[decision, draw=red] at(8,4) (c) {};
	\node[ecision,red] at(8,4) (cc) {\textbf{Labels} ($y_{i-\delta}$) available};
	\node[process] at(13,4) (d) {Fit \textbf{Classifier} with $(X_{i-\delta}, y_{i-\delta})$};
	\node[process] at(13,0) (e) {Return \textbf{predictions} ($y'_i$)};
	\node[start] at (16, 0) (stop) {$y'_i$};
			
	\draw[line] (start) -- (a);
	\draw[line,red] (a) --node[left,midway] {Y} (b);
	\draw[line] (b) -- (c);
	\draw[line,red] (c) --node[above,midway] {Y} (d);
	\draw[line] (d) -- (13,2) -- (e);
	\draw[line] (c) -- (8,0) --node[above, midway] {N} (e);
	\draw[line] (a) --node[above,midway] {N} (8,0) -- (c);
	\draw[line] (e) -- (stop);


  	\draw[lline] (-4,7) -- (20,7);
	\draw[lline] (-4,-3) -- (20,-3);
	\draw[lline] (-4,7) -- (-4,-3);
	\draw[lline] (20,7) -- (20,-3);

\end{tikzpicture}

%% file: frameworks/tr-p.tex
\begin{tikzpicture}


	\node[start] at (4,5) (start) {$X_i$};

	\node[decision, draw=red] at(7,5) (a) {};
	\node[ecision,red] at (7,5) (aa) {Unfulfilled request for \textbf{labels}};
	
	\node[decision, draw=red] at(7,0) (c) {};
	\node[ecision,red] at(7,0) (cc) {\textbf{Labels} ($y_{i-\delta}$) available};
	\node[process] at(11,0) (e) {Fit \textbf{Classifier} with ($X_{i-\delta},y_{i-\delta}$)};

	\node[decision, draw=red] at(15,0) (d) {};
	\node[ecision,red] at(15,0) (dd) {\textbf{Detection} $DD_p = f(X_{i-\delta},y_{i-\delta})$};
	
	\node[process] at(19,0) (f) {Request \textbf{labels} ($y_i$)};

	\node[decision, draw=red] at(19,5) (b) {};
	\node[ecision,red] at(19,5) (bb) {\textbf{Detection} $DD_p = f(X_{i})$};
	\node[process] at(23,0) (g) {Return \textbf{predictions} ($y'_i$)};

	\node[start] at (26,0) (stop) {$y'_i$};
	
	\draw[line] (start) -- (a);
	\draw[line,red] (a) --node[above,left] {Y} (c);
	\draw[line,red] (c) --node[above,midway] {Y} (e);
	\draw[line] (e) -- (d);
	\draw[line,red] (d) --node[above,midway] {Y} (f);
	\draw[line] (f) -- (g);

	\draw[line] (c) --node[left,midway] {N}  (7,-4) -- (23,-4) -- (g);
	\draw[line] (d) --node[left,midway] {N}  (15,-4) -- (23,-4) -- (g);
	\draw[line] (a) --node[left,above] {N}  (10,5) -- (b);

	\draw[line] (b) -- node[above,midway] {N} (23,5) -- (g);
	\draw[line,red] (b) -- node[left,midway] {Y} (f);
	\draw[line] (g) -- (stop);

	\draw[lline] (3,7) -- (27,7);
	\draw[lline] (3,-5) -- (27,-5);
	\draw[lline] (3,7) -- (3,-5);
	\draw[lline] (27,7) -- (27,-5);
 
\end{tikzpicture}